\documentclass[11pt,a4paper]{article}

\usepackage[a4paper, margin=2.5cm]{geometry}
\usepackage{amsmath, amssymb, amsfonts}
\usepackage{graphicx}
\usepackage{booktabs}
\usepackage{multirow}
\usepackage{array}
\usepackage{xcolor}
\usepackage{hyperref}
\usepackage{cleveref}
\usepackage{caption}
\usepackage{subcaption}
\usepackage{float}
\usepackage{microtype}
\usepackage{enumitem}
\usepackage{tabularx}
\usepackage{makecell}
\usepackage{colortbl}
\usepackage{setspace}
\usepackage{fancyhdr}
\usepackage{titlesec}
\usepackage{abstract}
\usepackage{natbib}
\usepackage{algorithm}
\usepackage{algpseudocode}
\usepackage{mathtools}
\usepackage{bm}

\definecolor{adultcol}{HTML}{2A9D8F}
\definecolor{childcol}{HTML}{2E86AB}
\definecolor{adocol}{HTML}{F4A261}
\definecolor{warnred}{HTML}{E76F51}
\definecolor{headblue}{HTML}{1F3864}
\definecolor{rowgray}{HTML}{F2F4F8}
\definecolor{goodgreen}{HTML}{D6EAD6}
\definecolor{badred}{HTML}{FAD4CC}
\definecolor{midyellow}{HTML}{FFF3CD}

\hypersetup{
  colorlinks=true,
  linkcolor=headblue,
  citecolor=adultcol,
  urlcolor=childcol,
  pdftitle={ASD-Bench},
  pdfauthor={Shubhankit Singh, Hassan Shaikh, Kuldeep Raghuwanshi, Keshav Bulia},
}

\titleformat{\section}{\large\bfseries\color{headblue}}{%
  \thesection}{1em}{}[\titlerule]
\titleformat{\subsection}{\normalsize\bfseries\color{headblue}}{\thesubsection}{1em}{}
\titleformat{\subsubsection}{\normalsize\bfseries\itshape}{\thesubsubsection}{1em}{}

\graphicspath{{figures/}}

\newcommand{\hap}{\textsc{hap}}

\newcommand{\wfn}{w_{\mathrm{FN}}}
\newcommand{\wfp}{w_{\mathrm{FP}}}

\newcolumntype{C}[1]{>{\centering\arraybackslash}p{#1}}
\newcolumntype{L}[1]{>{\raggedright\arraybackslash}p{#1}}
\newcolumntype{R}[1]{>{\raggedleft\arraybackslash}p{#1}}

\begin{document}

\begin{center}
{\LARGE\bfseries\color{headblue}
  ASD-Bench: A Four-Axis Comprehensive Benchmark \\[4pt] of AI Models for Autism Spectrum Disorder\\[4pt]
}
\vspace{1.2em}
{\large Shubhankit Singh$^{1,*}$ \quad Hassan Shaikh$^{1,2,\dagger}$ \quad Kuldeep Raghuwanshi$^{1,3,\dagger}$ \quad Keshav Bulia$^{1,2,\dagger}$}\\[0.4em]
{\normalsize $^{1}$Research Commons AI \quad $^{2}$IIT Bombay \quad $^{3}$IIT Delhi}\\[0.3em]
{\small $^{*}$Corresponding author \quad $^{\dagger}$Equal contribution}\\[0.3em]
{\small \texttt{shubhankitsingh@researchcommons.ai}} 


\vspace{0.6em}
{\small\textbf{Keywords:} Autism Spectrum Disorder, AQ-10, clinical AI, tabular benchmark, AI/ML models}
\end{center}

\hrule
\vspace{1.5em}

\begin{abstract}
\noindent
Automated ASD screening tools remain limited by single-architecture evaluations,
axis-restricted assessment, and near-exclusive focus on adult cohorts, obscuring
age-specific diagnostic patterns critical for early intervention. We introduce
\textbf{ASD-Bench}, a systematic tabular benchmark evaluating ML, deep learning,
and foundation model configurations across three age
cohorts (children 1--11\,yr, adolescents 12--16\,yr, adults 17--64\,yr) on four
axes: predictive performance, calibration, interpretability, and
adversarial robustness. Applied to a curated v3 dataset of 4{,}068 AQ-10
records, our benchmark spans classical models (XGBoost, AdaBoost,
Random Forest, Logistic Regression), neural networks (MLP), deep tabular
transformers (TabNet, TabTransformer, FT-Transformer), and TabPFN~v2.
We introduce the \textbf{Heuristic Aggregate
Penalty (HAP)}: a cost-sensitive metric penalising false negatives more
heavily and incorporating cross-validation variance for
deployment stability. Adult classification yields high performance
(10/17 models achieve perfect F1 and AUC), while
adolescents present a harder task (F1 ceiling 0.837 vs.\ 0.915 for
children). Feature hierarchies shift across cohorts: A9 (social
motivation) dominates for children, A5 (pattern recognition)
leads for adolescents, and adults exhibit a flatter importance profile
consistent with developmental social masking.
Accuracy and calibration are dissociated: AdaBoost achieves F1\,=\,1.000 on
adults with ECE\,=\,0.302, confirming single-metric evaluation is
insufficient for clinical AI. Cohort-specific deployment recommendations
are provided. All findings should be interpreted
as proof-of-concept evidence on questionnaire-derived labels rather than
clinically validated diagnostic performance.
\end{abstract}

\vspace{1em}
\hrule

\section{Introduction}
\label{sec:intro}

According to the WHO Report 2021, one in every 127 people is affected by Autism Spectrum Disorder (ASD)~\cite{who2021}, a lifelong neurodevelopmental condition characterised by persistent challenges in social communication, restricted and repetitive behaviours, and atypical sensory processing~\cite{dsm5}. The heterogeneity of symptom presentation is captured by the term \emph{spectrum}, which spans mild social difficulties to severe communicative impairment. Over the past two decades, ASD prevalence has grown significantly~\cite{lundstrom2015}, placing pressure on healthcare systems already burdened by specialist shortages and high assessment costs~\cite{baird2006}. Since timely intervention substantially improves cognitive, social, and behavioural outcomes~\cite{howlin2004}, early identification is clinically critical and need to addressed by the Autism-Spectrum Quotient 10-item (AQ-10) instrument, which enables rapid, low-cost first-pass assessment suitable for primary care.

Prior work on automated ASD screening spans three threads. Classical machine learning methods: Support Vector Machines, Random Forests, and $k$-Nearest Neighbours applied to AQ-10 questionnaire data~\cite{thabtah2019,thabtah2017,allison2012} which remain competitive on tabular inputs, while deep learning extensions to neuroimaging~\cite{Bayram2021,heinsfeld2017,taban2021,kong2019,li2022}, eye-tracking~\cite{fang2020}, and video~\cite{tariq2018} achieve strong within dataset accuracy but require specialised acquisition hardware unavailable in routine clinical settings. In parallel, modern deep tabular architectures have emerged, TabNet~\cite{arik2021} with sequential attention and instance-wise feature masks, TabTransformer~\cite{huang2020} and FT-Transformer~\cite{gorishniy2021} applying self-attention over feature embeddings, and TabPFN~v2~\cite{hollmann2023} performing in-context learning via a prior-data fitted network, yet no ASD study has systematically compared them in a unified, highest unification have done in ~\cite{nithya2025autism} which did the Lime interprretability and heurtisc categorization to propose the educational plan for students with no caliberation measures. Finally, clinical deployment demands evaluation beyond accuracy: calibration~\cite{guo2017}, robustness to distributional shift~\cite{degrave2021}, and interpretability via SHAP~\cite{lundberg2017} and LIME~\cite{ribeiro2016} are all essential, yet no existing ASD study formalises a composite metric that penalises misclassification asymmetrically and rewards stability across these axes.

These threads expose three concrete gaps. Most studies rely on \emph{single-architecture evaluation}, comparing only one or two model families and omitting recent deep tabular learners on behavioural questionnaire data; they offer \emph{axis-limited assessment}, reporting accuracy or F1 alone while neglecting calibration, interpretability, and robustness; and they typically focus on a \emph{single age cohort} almost exclusively adults or children masking age-specific diagnostic patterns. Adolescents in particular remain under-studied despite exhibiting distinct feature hierarchies and a harder classification task (F1 ceiling of 0.837 vs 0.915 for children).

This study addresses these gaps through four contributions. First, we construct Dataset~v3 by combining UCI AQ-10 data (v1) with a supplementary source (v2), yielding 4,068 records across three cohorts after deduplication and quality control. Second, we conduct a 17-model systematic benchmark encompassing classical ensembles, neural networks, and deep tabular transformers each with baseline and hyperparameter-tuned variants plus a foundation model. Third, we establish a four-axis evaluation framework covering predictive performance, calibration, interpretability, and adversarial robustness. Finally, we introduce the \hap{} (Heuristic Aggregate Penalty) metric: a clinically motivated composite score incorporating asymmetric FN/FP penalties and a cross-fold variance term.
\section{Dataset and Preprocessing}
\label{sec:data}


Our v3 dataset (Table~\ref{tab:dataset}) integrates two sources. The primary source (v1) is the UCI Machine Learning Repository ASD Screening dataset~\cite{thabtah2017}, containing AQ-10 questionnaire responses (Table~\ref{tab:aq10}) and demographics for adult, adolescent, and child participants. The secondary source (v2) is an additional ASD screening dataset from the University of Arkansas, Department of Computer Science~\cite{afarinbargrizan2024asd}, providing supplementary records with an identical AQ-10 structure. Although both data versions were collected through the ASDTest app, we identified distinct data points across sources and combined them through a two-stage preprocessing pipeline: the first stage performed deduplication by removing records shared between v1 and v2, while the second stage carried out data cleaning and quality control on a per-cohort basis (refer to the dataset link in the \emph{Data and Code Availability} section). After processing, the final dataset contains \textbf{4{,}068 records} across three age cohorts.

The merged v3 corpus is near-balanced overall (52.5\% ASD-positive), with a gender distribution of 67.6\% male and 32.4\% female. Ethnicity composition is: White European 30.4\%, Asian 27.8\%, Middle Eastern 17.7\%, South Asian 9.0\%, and Black 4.1\%. We note that fairness analysis of the ASD label across these sub-categories is beyond the scope of the present study.

\begin{table}[H]
\centering
\caption{ASD-Bench v3 dataset composition and quality summary (post all cleaning steps).}
\label{tab:dataset}
\renewcommand{\arraystretch}{1.25}
\begin{tabular}{L{2.4cm} C{1.8cm} C{1.8cm} C{2.2cm} C{3.0cm}}
\toprule
\textbf{Cohort} & \textbf{Final Records} & \textbf{Age Range} &
\textbf{Removed Duplicates}$^a$ & \textbf{ASD YES\,/\,NO (post-clean)} \\
\midrule
Child       & 2{,}514 & 1--11\,yr  & 6 (2.1\%)   & $\approx$60\%\,/\,40\% \\
Adolescent  & 818     & 12--16\,yr & 0 (0.0\%)   & $\approx$53\%\,/\,47\% \\
Adult       & 736     & 17--64\,yr & 380 (54.0\%) & $\approx$26\%\,/\,74\% \\
\midrule
\textbf{Combined} & \textbf{4{,}068} & 1--64\,yr &
\textbf{386 total} & 52.5\%\,/\,47.5\% \\
\bottomrule
\end{tabular}
\\[2pt]
\end{table}

\begin{table}[H]
\centering
\caption{AQ-10 questionnaire items encoded as binary features A1--A10.
         $\star$\,=\,reverse-scored. A threshold of $\geq 6$ suggests elevated
         autistic traits}
\label{tab:aq10}
\renewcommand{\arraystretch}{1.2}
\begin{tabular}{C{0.9cm} L{8.5cm} C{3.0cm}}
\toprule
\textbf{ID} & \textbf{AQ-10 Question} & \textbf{Scoring} \\
\midrule
A1  & I prefer to do things the same way over and over again.            & 1\,=\,Agree \\
A2  & I find it hard to make small talk.                                  & 1\,=\,Agree \\
A3  & I would rather go to a party than a museum.                         & 1\,=\,Disagree$^\star$ \\
A4  & I get highly upset if my routine is disrupted.                      & 1\,=\,Agree \\
A5  & I notice patterns in things all the time.                           & 1\,=\,Agree \\
A6  & I frequently don't know how to keep a conversation going.           & 1\,=\,Agree \\
A7  & When reading a story, I find it hard to understand characters' intentions. & 1\,=\,Agree \\
A8  & I find it easy to work out what someone is thinking by looking at their face. & 1\,=\,Disagree$^\star$ \\
A9  & I enjoy social chit-chat.                                           & 1\,=\,Disagree$^\star$ \\
A10 & I find it easy to understand what others are thinking.              & 1\,=\,Disagree$^\star$ \\
\bottomrule
\end{tabular}
\end{table}

\section{Methodology}
\label{sec:method}

\subsection{Models and Configurations}

We evaluate four categories of models, each trained in both a default (baseline) and hyperparameter-tuned configuration via \texttt{GridSearchCV} unless stated otherwise.

\textbf{Classical models.}
We include Logistic Regression~\cite{cox1958regression}, Random Forest~\cite{breiman2001random}, AdaBoost~\cite{freund1997}, and XGBoost~\cite{chen2016} as well-established tabular baselines. A fully-connected Multi-Layer Perceptron (MLP) implemented in scikit-learn is included as a shallow neural baseline.

\textbf{Deep tabular architectures.}
Three attention-based architectures are evaluated. TabNet~\cite{arik2021} uses sequential entmax masking (batch size 64, virtual batch 32, early-stopping patience 10). TabTransformer~\cite{huang2020} applies column-wise self-attention with $d_{\mathrm{model}}=32$, 3 layers, and 8 heads. FT-Transformer~\cite{gorishniy2021} uses a CLS-token classification head with $d_{\mathrm{model}}=32$. To quantify predictive uncertainty, Monte Carlo Dropout ($T=20$ stochastic forward passes) is applied to all three transformer variants at inference time.

\textbf{Foundation model.}
TabPFN~v2~\cite{hollmann2023} is a prior-data fitted network pre-trained
on synthetic task distributions that performs in-context learning at
inference time: the target training set is passed as context without any
gradient-based fine-tuning or hyperparameter sweep. We use 8 estimators
with CPU inference. Because TabPFN receives no task-specific optimisation
while all other models undergo full hyperparameter search, any direct
ranking comparison is inherently asymmetric and favours the tuned models;
TabPFN results should be read as a lower bound on foundation-model
performance for this task.

\subsection{Training Protocol}

All models were trained independently per cohort using a fixed random seed
(42). A stratified 80/20 train--test split was applied globally before any
training; the held-out 20\% serves as the final evaluation set for all
reported metrics with conducting 5-fold cross validation on the train dataset. Deep models were optimised with Adam for up to 100 epochs,
with early stopping (patience\,=\,10) within each fold.

\subsection{Four-Axis Evaluation Framework}

\subsubsection{Axis 1 --- Predictive Performance}
Accuracy, precision, recall, F1-score, and AUC-ROC on the held-out test set per
cohort. These standard classification metrics quantify discriminative performance per cohort.

\subsubsection{Axis 2 --- Calibration / Uncertainty Estimation}
Expected Calibration Error (ECE), Brier score, mean
confidence, confidence standard deviation, and mean prediction entropy.
MC Dropout ($T=20$) is applied to transformer variants for epistemic uncertainty estimation.

\begin{table}[h]
\centering
\renewcommand{\arraystretch}{1.4}
\begin{tabular}{|l|l|c|c|l|}
\hline
\textbf{Metric} & \textbf{Formula} & \textbf{Range} & \textbf{Preferred} & \textbf{Measures} \\
\hline
Mean Confidence
    & $\dfrac{1}{n}\displaystyle\sum \max(p_i)$
    & $[0,1]$ & Higher & Average certainty \\
\hline
Std Confidence
    & $\sqrt{\mathrm{Var}(\max(p_i))}$
    & $[0,0.5]$ & Lower & Consistency \\
\hline
Entropy
    & $-\displaystyle\sum p\log(p)$
    & $[0,1]$ & Lower & Uncertainty \\
\hline
Brier Score
    & $\dfrac{1}{n}\displaystyle\sum (p_i - y_i)^2$
    & $[0,1]$ & Lower & Accuracy + Calibration \\
\hline
ECE
    & $\displaystyle\sum \frac{|B_m|}{n}\,|\,\mathrm{acc} - \mathrm{conf}\,|$
    & $[0,1]$ & Lower & Calibration quality \\
\hline
\end{tabular}
\caption{Calibration and uncertainty metrics used in 2nd Axis evaluation.}
\label{tab:axis2_metrics}
\end{table}

\subsubsection{Axis 3 --- Interpretability}
SHAP TreeExplainer (XGBoost, RF), SHAP DeepExplainer/GradientExplainer (MLP),
permutation importance (AdaBoost, TabTransformer, FT-Transformer and TabPFN), Logistic Regression coefficients, and
TabNet built-in feature masks. A \emph{consensus importance} is computed by
averaging normalised scores across all 17 applicable models, with tool selection determined by model architecture (gradient availability for neural networks, tree structure for ensembles).

\subsubsection{Axis 4 --- Robustness Testing}
Three adversarial perturbation protocols on the test set:
\begin{enumerate}[leftmargin=1.5em]
  \item \textbf{Feature Flip:} randomly flip $k \in \{10\%, 20\%, 30\%\}$ of
        binary feature values.
  \item \textbf{Gaussian Noise:} add/subtract $\varepsilon \sim \mathcal{N}(0, \sigma^2)$,
        $\sigma \in \{0.1, 0.2, 0.3\}$, clipped to $[0,1]$.
  \item \textbf{Feature Removal:} zero-out the top-$k \in \{1,2,3\}$ most
        important features.
\end{enumerate}
Composite robustness score: $R = 1 - \overline{\Delta\text{acc}}$ across all
perturbation levels. This axis assesses each model's capacity to maintain predictions under noisy or erroneous input conditions (whether due to data entry errors or intentional misreporting). Additionally, given that ASD is better characterised as a construct space than a binary category, robustness testing reveals how models handle borderline cases through Gaussian noise degradation.

\subsection{HAP: Heuristic Aggregate Penalty}
\label{sec:hap}
 
Standard accuracy and F1 metrics treat false positives~(FP) and false
negatives~(FN) symmetrically. In clinical ASD screening, a false negative (missing a
genuine case) deprives a child of early intervention, while a false positive
leads only to an unnecessary follow-up referral. We formalise
this asymmetry through the \textbf{Heuristic Aggregate Penalty}~(\hap{}).
 
\paragraph{Cost function.}
Given a confusion matrix $\{\mathrm{TP},\mathrm{TN},\mathrm{FP},\mathrm{FN}\}$
and penalty weights $\wfp=2$, $\wfn=10$
(with $w_{\mathrm{TP}}=w_{\mathrm{TN}}=0$),
the per-fold weighted cost is
\begin{equation}
  \mathcal{C}_{k} \;=\;
  \frac{\wfp\cdot\mathrm{FP}_{k} + \wfn\cdot\mathrm{FN}_{k}}{N_{k}},
  \label{eq:cost}
\end{equation}
where $N_{k}$ is the fold sample count.
The ratio $\wfn:\wfp = 5:1$ is \emph{conservative} relative to the
cost effectiveness; a sensitivity
analysis over the range $\wfn:\wfp \in [1,\,20]$ confirms that model
rankings are stable across this sweep ($W = 0.995$, $0.996$, and $1.000$ for the adolescent, child, 
and adult cohorts respectively,
Appendix), so conclusions are not driven by the specific choice.
 
\paragraph{Stability-penalised aggregation.}
\hap{} aggregates $\mathcal{C}_{k}$ over stratified $K$\,=\,5-fold
cross-validation with an explicit variance penalty:
\begin{equation}
  \hap \;=\;
  \underbrace{\frac{1}{K}\sum_{k=1}^{K}\mathcal{C}_{k}}_{\text{mean cost}}
  \;+\;
  \lambda\cdot
  \underbrace{\mathrm{Var}_{k}\!\left(\mathcal{C}_{k}\right)}_{\text{instability}},
  \label{eq:hap}
\end{equation}
rewarding models that are consistent across data partitions---a
critical property for multi-site clinical deployment.
 
\paragraph{Principled selection of $\lambda$.}
Rather than fixing $\lambda$ arbitrarily, we derive it from a
discrimination signal-to-noise criterion.
Each model $i$ traces a linear score trajectory
$\mathrm{HAP}_{i}(\lambda)=\mu_{i}+\lambda\sigma_{i}^{2}$
in $\lambda$-space, where $\mu_{i}$ and $\sigma_{i}^{2}$ are its
cross-validated mean cost and variance.
Define the \emph{inter-model separation}
$S(\lambda)=\binom{N}{2}^{-1}\sum_{i<j}|\mathrm{HAP}_{i}-\mathrm{HAP}_{j}|$
and the discrimination SNR:
\begin{equation}
  \mathrm{SNR}(\lambda)
  \;=\;
  \frac{S(\lambda)}{\sqrt{\mathrm{Var}_{i}\!\left[\mathrm{HAP}_{i}(\lambda)\right]}}.
  \label{eq:snr}
\end{equation}
Because $S(\lambda)\approx A+B\lambda$ (linear) while
$\mathrm{Var}_{i}[\mathrm{HAP}_{i}(\lambda)]= C + D\lambda + E\lambda^{2}$
(quadratic, with $C=\mathrm{Var}(\mu)$,
$D=2\,\mathrm{Cov}(\mu,\sigma^{2})$,
$E=\mathrm{Var}(\sigma^{2})$),
$\mathrm{SNR}(\lambda)$ is unimodal and attains its maximum at:
\begin{equation}
  \lambda^{*}
  \;=\;
  \frac{AE + \sqrt{A^{2}E^{2}+BE(BC-AD)}}{BE},
  \label{eq:lambdastar}
\end{equation}
verified by $d^{2}\mathrm{SNR}/d\lambda^{2}\big|_{\lambda^{*}}<0$.
Beyond $\lambda^{*}$, pairwise rank crossovers
$\lambda_{ij}^{\times}=(\mu_{j}-\mu_{i})/(\sigma_{i}^{2}-\sigma_{j}^{2})$
begin to proliferate, causing rankings to reflect variance differences
rather than mean performance---a dispersive regime analogous to
gain-induced instability in linear control systems.
$\lambda=1.0$ lies within the stable discrimination lobe,
achieving $98.6$--$99.4\%$ of the theoretical maximum SNR while remaining
an interpretable unit-consistent choice ($\mu$ and $\sigma^{2}$ share
the same cost scale).

\section{Results}
\label{sec:results}

\subsection{Predictive Performance}
\label{sec:perf}

\begin{figure}[H]
  \centering
  \includegraphics[width=\textwidth]{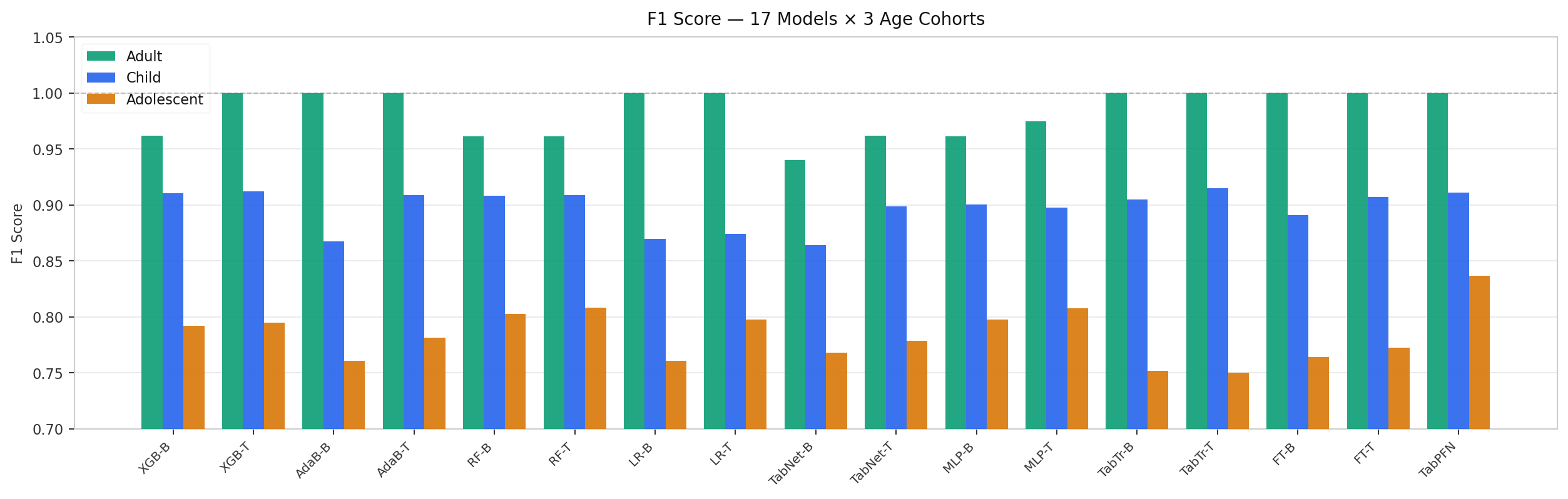}
  \caption{F1 Score for all 17 models across three age cohorts
           (\textcolor{adultcol}{\textbf{Adult}},
            \textcolor{childcol}{\textbf{Child}},
            \textcolor{adocol}{\textbf{Adolescent}}). Adults achieve perfect F1 for 10 of 17
           models; child F1 peaks at 0.915, adolescent at 0.837.}
  \label{fig:f1}
\end{figure}

\begin{figure}[H]
  \centering
  \includegraphics[width=\textwidth]{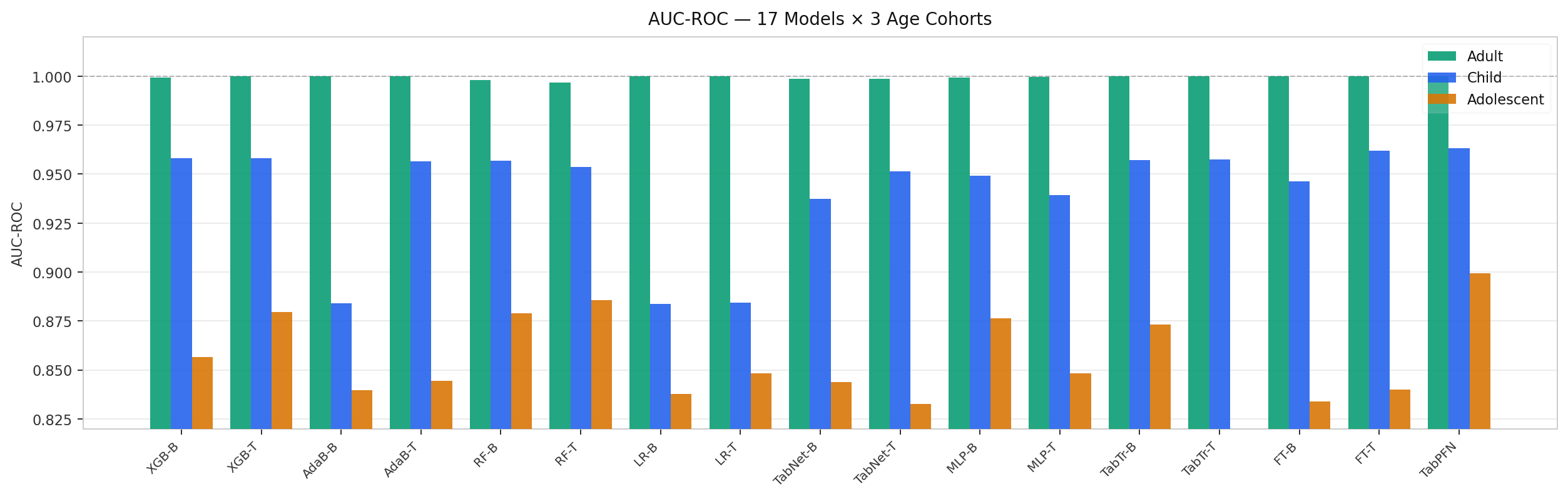}
  \caption{AUC-ROC for all 17 models across three age cohorts.
           Adults: 11 of 17 models at AUC\,=\,1.000.
           TabPFN achieves the highest child AUC (0.963) and adolescent AUC (0.900).}
  \label{fig:auc}
\end{figure}

\begin{figure}[H]
  \centering
  \includegraphics[width=\textwidth]{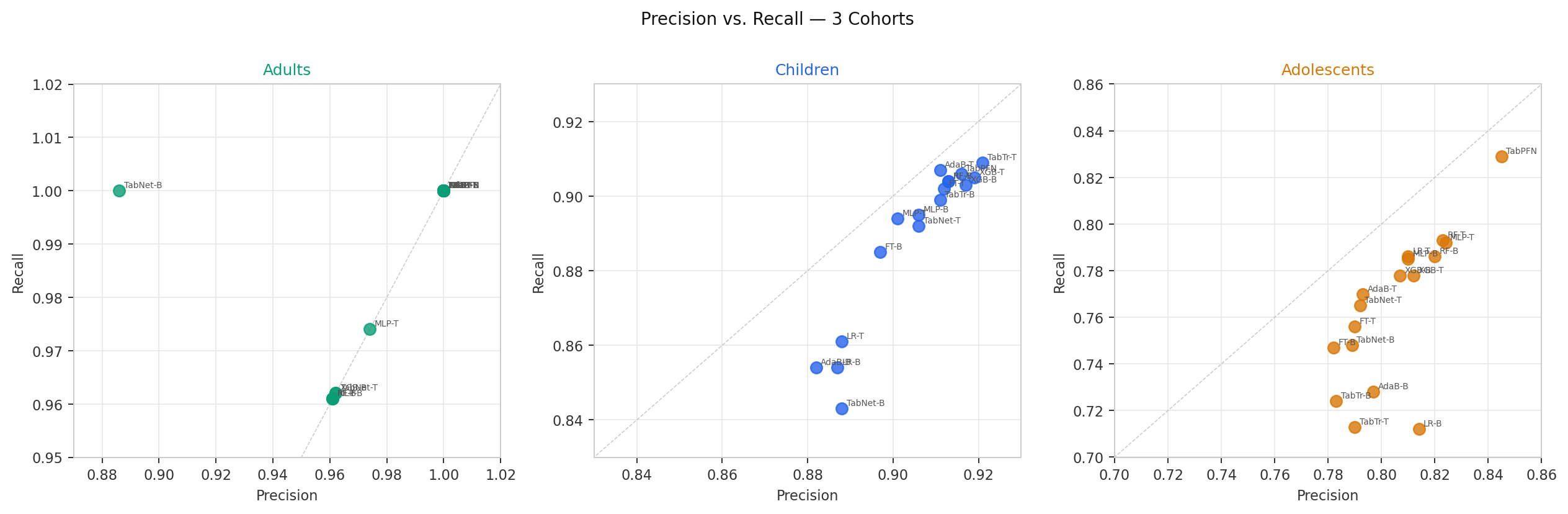}
  \caption{Precision vs.\ Recall scatter per cohort. Diagonal dashed line:
           equal precision\,/\,recall. TabNet Baseline achieves recall\,=\,1.000
           on adults at the cost of precision\,=\,0.886.}
  \label{fig:pr_scatter}
\end{figure}

Figures~\ref{fig:f1}--\ref{fig:pr_scatter} present F1, AUC-ROC, and
precision--recall distributions across all three cohorts.

\paragraph{Adults.} Ten of 17 models achieve F1\,=\,1.000 and AUC\,=\,1.000,
confirming near-perfect separability of the adult AQ-10 feature space. XGBoost
Baseline (F1\,=\,0.962) and TabNet Baseline (F1\,=\,0.940) are the only
notable underperformers; the latter achieves recall\,=\,1.000 at
precision\,=\,0.886.

\paragraph{Adolescents.} F1 ranges 0.750--0.837, a 7.8 percentage-point gap
below the child ceiling. TabPFN leads with F1\,=\,0.837 and AUC\,=\,0.900.
TabTransformer Tuned (the best-performing child model) achieves only F1\,=\,0.750
and AUC\,=\,0.816 on adolescents, confirming that the adolescent cohort presents a
harder classification task. Random Forest Tuned achieves the
second-highest adolescent F1 (0.808), suggesting tree-ensemble stability is more
valuable when the feature-label mapping is weaker.

\paragraph{Children.} F1 ranges 0.864--0.915. TabTransformer Tuned leads on F1
(0.915), TabPFN achieves the highest AUC (0.963). Simpler models
(AdaBoost Baseline, Logistic Regression) form the lower tier
(F1\,$\approx$\,0.867--0.870).

\begin{table}[H]
\centering
\caption{F1-score and AUC-ROC for all 17 models across three cohorts.
         $\star$\,=\,cohort best.}
\label{tab:f1_auc}
\renewcommand{\arraystretch}{1.2}
\setlength{\tabcolsep}{5pt}
\small
\begin{tabular}{L{3.2cm} C{1.3cm} C{1.3cm} C{1.3cm} C{1.3cm} C{1.3cm} C{1.3cm}}
\toprule
\textbf{Model} &
\textbf{Ch.\ F1} & \textbf{Ch.\ AUC} &
\textbf{Ado.\ F1} & \textbf{Ado.\ AUC} &
\textbf{Ad.\ F1} & \textbf{Ad.\ AUC} \\
\midrule
XGBoost Baseline      & 0.910 & 0.958 & 0.792 & 0.857 & 0.962 & 0.999 \\
XGBoost Tuned         & 0.912 & 0.958 & 0.795 & 0.880 & 1.000 & 1.000 \\
AdaBoost Baseline     & 0.867 & 0.884 & 0.761 & 0.840 & 1.000 & 1.000 \\
AdaBoost Tuned        & 0.909 & 0.956 & 0.782 & 0.844 & 1.000 & 1.000 \\
RF Baseline           & 0.908 & 0.957 & 0.803 & 0.879 & 0.961 & 0.998 \\
RF Tuned              & 0.909 & 0.954 & 0.808 & 0.886 & 0.961 & 0.997 \\
LR Baseline           & 0.870 & 0.884 & 0.761 & 0.838 & 1.000 & 1.000 \\
LR Tuned              & 0.874 & 0.884 & 0.798 & 0.848 & 1.000 & 1.000 \\
TabNet Baseline       & 0.864 & 0.938 & 0.768 & 0.843 & 0.940 & 0.999 \\
TabNet Tuned          & 0.899 & 0.951 & 0.779 & 0.833 & 0.962 & 0.999 \\
MLP Baseline          & 0.901 & 0.949 & 0.797 & 0.876 & 0.961 & 0.999 \\
MLP Tuned             & 0.897 & 0.939 & 0.808 & 0.848 & 0.974 & 1.000 \\
TabTransformer Base   & 0.905 & 0.957 & 0.752 & 0.873 & 1.000 & 1.000 \\
TabTransformer Tuned  & 0.915$^\star$ & 0.957 & 0.750 & 0.816 & 1.000 & 1.000 \\
FT-Transformer Base   & 0.891 & 0.946 & 0.764 & 0.834 & 1.000 & 1.000 \\
FT-Transformer Tuned  & 0.907 & 0.962 & 0.772 & 0.840 & 1.000 & 1.000 \\
TabPFN v2             & 0.911 & 0.963$^\star$ & 0.837$^\star$ & 0.900$^\star$ & 1.000 & 1.000 \\
\bottomrule
\end{tabular}
\end{table}

\subsection{Calibration Quality}
\label{sec:calib}

\begin{figure}[H]
  \centering
  \includegraphics[width=0.92\textwidth]{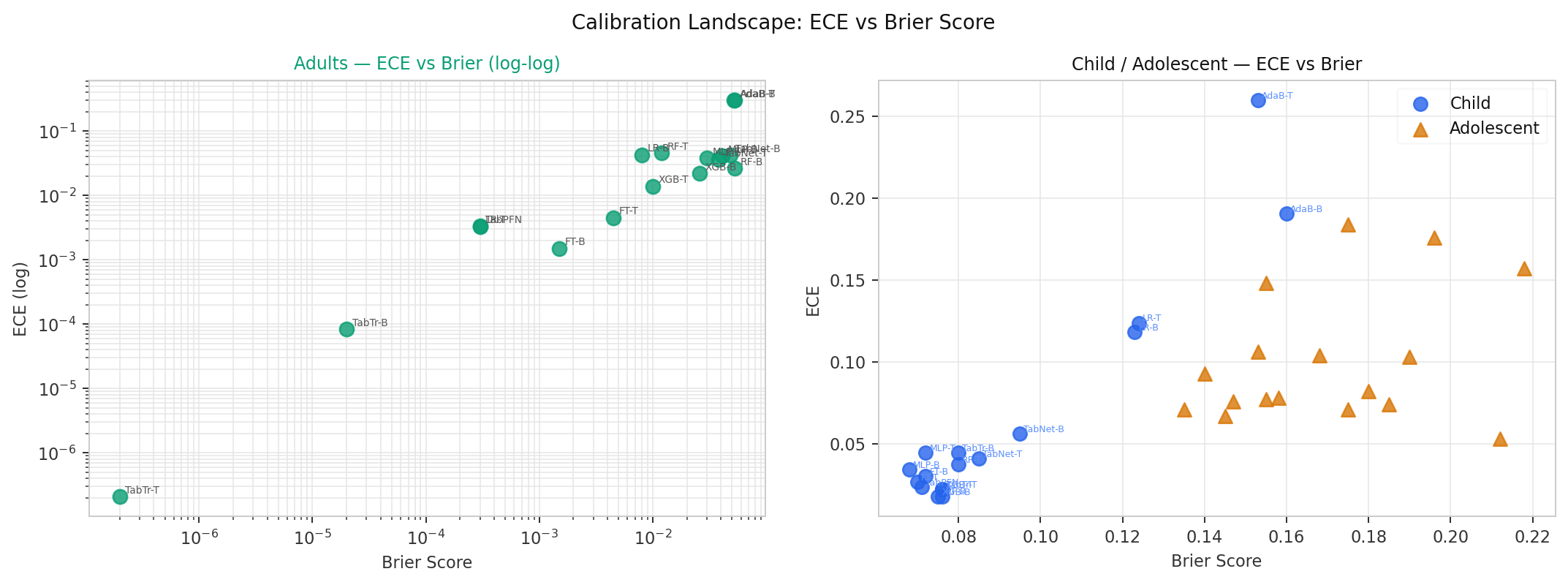}
  \caption{ECE vs.\ Brier Score per cohort (bottom-left\,=\,ideal).
           Adult on log-log scale; child/adolescent on linear scale.
           AdaBoost is a clear outlier in both panels.}
  \label{fig:ece_brier}
\end{figure}

\begin{table}[H]
\centering
\caption{%
  \textbf{(5a)} ECE for all models across
  age cohorts. $\star$\,=\,cohort best;
  $\dagger$\,=\,critical (ECE\,$>\,0.12$).
}
\label{tab:ece_a}
\renewcommand{\arraystretch}{1.2}
\setlength{\tabcolsep}{5pt}
\small
\begin{tabular}{L{4.5cm} C{2.4cm} C{2.0cm} C{2.0cm}}
\toprule
\textbf{Model}
  & \textbf{ECE --- Adult}
  & \textbf{ECE --- Child}
  & \textbf{ECE --- Adolescent} \\
\midrule
XGBoost Baseline
  & 0.022
  & $0.018^{\star}$
  & 0.078 \\
XGBoost Tuned
  & 0.014
  & 0.022
  & 0.106 \\
AdaBoost Baseline
  & $0.302^{\dagger}$
  & $0.190^{\dagger}$
  & $0.157^{\dagger}$ \\
AdaBoost Tuned
  & $0.303^{\dagger}$
  & $0.260^{\dagger}$
  & $0.176^{\dagger}$ \\
RF Baseline
  & 0.026
  & $0.018^{\star}$
  & $0.067^{\star}$ \\
RF Tuned
  & 0.046
  & 0.037
  & 0.093 \\
LR Baseline
  & 0.043
  & $0.118$
  & $0.053$ \\
LR Tuned
  & 0.003
  & $0.124^{\dagger}$
  & $0.184^{\dagger}$ \\
TabNet Baseline
  & 0.042
  & 0.056
  & 0.077 \\
TabNet Tuned
  & 0.036
  & 0.041
  & 0.104 \\
MLP Baseline
  & 0.041
  & 0.034
  & 0.076 \\
MLP Tuned
  & 0.038
  & 0.044
  & $0.148^{\dagger}$ \\
TabTransformer Base
  & $8.3\!\times\!10^{-5}$
  & 0.044
  & 0.082 \\
TabTransformer Tuned
  & $2.1\!\times\!10^{-7\,\star}$
  & 0.022
  & 0.103 \\
FT-Transformer Base
  & $1.5\!\times\!10^{-3}$
  & 0.030
  & 0.074 \\
FT-Transformer Tuned
  & $4.5\!\times\!10^{-3}$
  & 0.027
  & $0.071$ \\
TabPFN v2
  & $3.3\!\times\!10^{-3}$
  & 0.024
  & 0.071 \\
\bottomrule
\end{tabular}
\end{table}

The threshold of ECE $> 0.12$ was chosen empirically, lying approximately one standard deviation above the mean of the observed ECE distribution and coinciding with a natural gap between the well-calibrated cluster (ECE $\leq 0.106$) and poorly calibrated models (ECE $\geq 0.118$).
 
\begin{table}[H]
\centering
\caption{%
  \textbf{(5b)} Brier Score and Mean Confidence for \textbf{calibration-safe
  models only} (no ECE\,$>\,0.12$ in any cohort from Table~\ref{tab:ece_a};
  4 models excluded: AdaBoost Baseline/Tuned, LR Tuned, MLP Tuned;
  LR Baseline borderline at 0.118 and excluded as a precaution).
  Brier Score lower is better; Mean Confidence closer to 1.0 indicates
  decisive predictions.
  $\star$\,=\,cohort best Brier.
}
\label{tab:brier_b}
\renewcommand{\arraystretch}{1.2}
\setlength{\tabcolsep}{4.5pt}
\small
\begin{tabular}{L{3.4cm} C{1.6cm} C{1.6cm} C{1.6cm} C{1.6cm} C{1.6cm} C{1.6cm}}
\toprule
\textbf{Model}
  & \textbf{Brier Adult}
  & \textbf{Brier Child}
  & \textbf{Brier Adolescent}
  & \textbf{Conf. Adult}
  & \textbf{Conf. Child}
  & \textbf{Conf. Adolescent} \\
\midrule
XGBoost Baseline
  & 0.014
  & 0.075
  & 0.128
  & 0.965
  & 0.843
  & 0.782 \\
XGBoost Tuned
  & $0.000^{\star}$
  & 0.074
  & 0.131
  & 1.000
  & 0.848
  & 0.794 \\
\midrule
RF Baseline
  & 0.026
  & 0.076
  & 0.124
  & 0.948
  & 0.841
  & 0.800 \\
RF Tuned
  & 0.026
  & 0.077
  & $0.118^{\star}$
  & 0.947
  & 0.839
  & 0.806 \\
\midrule
TabNet Baseline
  & 0.042
  & 0.092
  & 0.133
  & 0.939
  & 0.809
  & 0.779 \\
TabNet Tuned
  & 0.027
  & 0.082
  & 0.139
  & 0.960
  & 0.820
  & 0.768 \\
MLP Baseline
  & 0.028
  & 0.082
  & 0.126
  & 0.951
  & 0.820
  & 0.793 \\
\midrule
TabTransformer Base
  & $0.000^{\star}$
  & 0.080
  & 0.137
  & 1.000
  & 0.827
  & 0.773 \\
TabTransformer Tuned
  & $0.000^{\star}$
  & 0.076
  & 0.140
  & 1.000
  & 0.840
  & 0.768 \\
FT-Transformer Base
  & $0.000^{\star}$
  & 0.078
  & 0.136
  & 1.000
  & 0.833
  & 0.774 \\
FT-Transformer Tuned
  & $0.000^{\star}$
  & 0.077
  & 0.133
  & 1.000
  & 0.836
  & 0.779 \\
\midrule
TabPFN v2
  & $0.000^{\star}$
  & $0.071^{\star}$
  & $0.113^{\star}$
  & 1.000
  & 0.854
  & 0.820 \\
\bottomrule
\end{tabular}
\end{table}

A critical finding is that high predictive accuracy does not imply good calibration:
AdaBoost achieves F1\,=\,1.000 on adults with ECE\,=\,0.302, rendering its probability
outputs untrustworthy for clinical threshold-based decisions, while the adult cohort
overall achieves the strongest calibration, with five models attaining ECE\,$<\,0.005$
and TabTransformer Tuned reaching $2.1\!\times\!10^{-7}$. The adolescent cohort exhibits
systematically weaker calibration than children: the best calibration-safe adolescent ECE
(RF Baseline: 0.067) is nearly 4$\times$ worse than the best child ECE (XGBoost Baseline:
0.018), and LR Tuned degrades from ECE\,=\,0.053 to 0.184 after tuning, suggesting
overfitting to adolescent noise.

\subsection{Interpretability and Feature Attribution}
\label{sec:interp}

\begin{figure}[H]
  \centering
  \includegraphics[width=\textwidth]{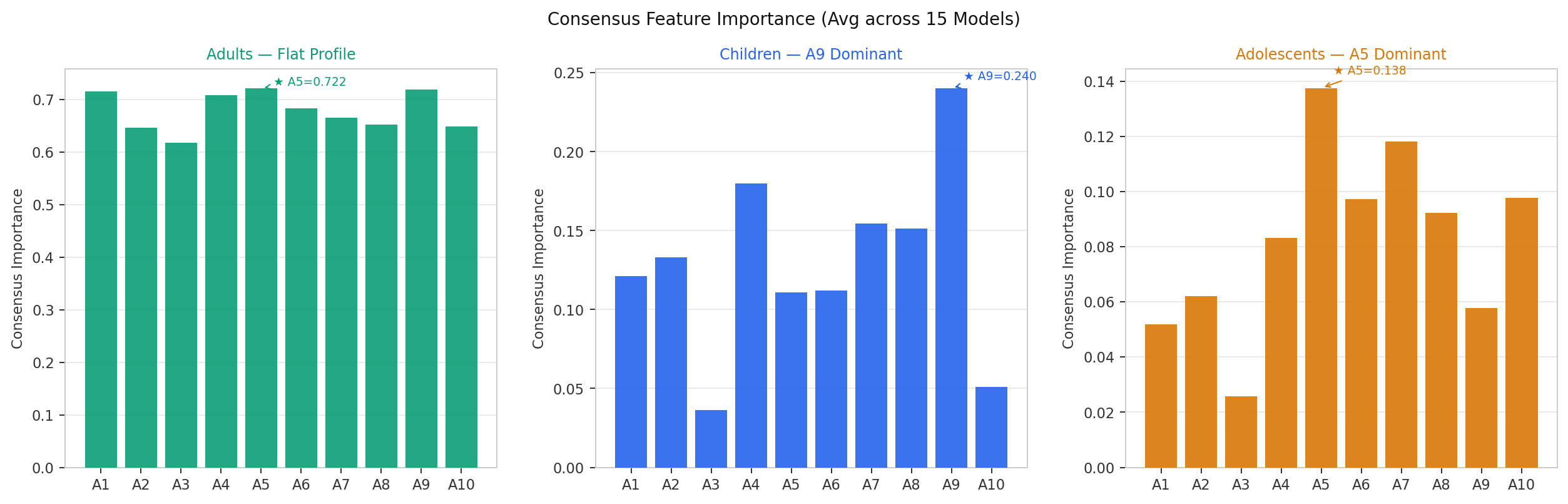}
  \caption{Consensus feature importance (averaged across 17 applicable models,
           normalised) per cohort. $\bigstar$\,=\,top-ranked feature per cohort.
           Note distinct hierarchies: A9 dominates children; A5 leads
           adolescents; adults show a flat multi-feature profile.}
  \label{fig:feat_imp}
\end{figure}

\begin{figure}[H]
  \centering
  \includegraphics[width=0.88\textwidth]{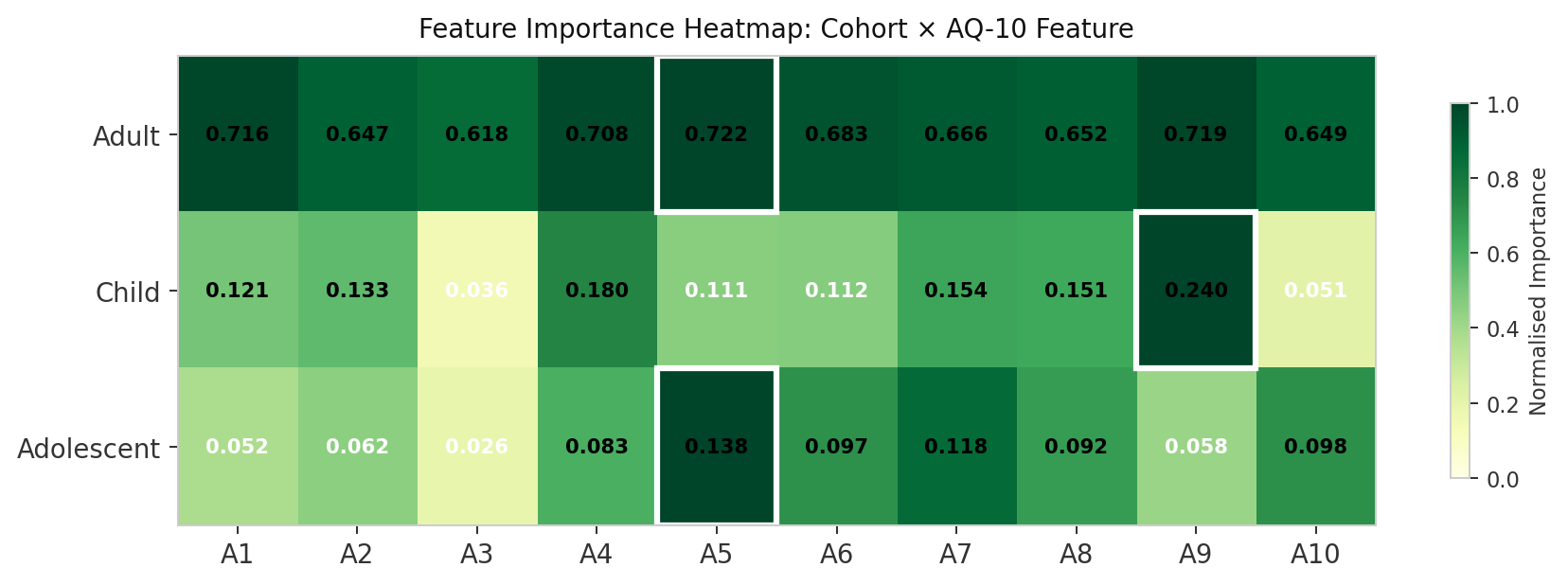}
  \caption{Feature importance heatmap: cohort $\times$ AQ-10 feature.
           Bold values\,=\,cohort maximum. The three cohorts show
           distinct importance profiles.}
  \label{fig:fi_heatmap}
\end{figure}

\begin{table}[H]
\centering
\caption{Consensus feature importance by cohort ($\star$\,=\,cohort maximum).}
\label{tab:feat}
\renewcommand{\arraystretch}{1.2}
\small
\begin{tabular}{C{1.0cm} C{1.8cm} C{1.8cm} C{1.8cm} L{5.5cm}}
\toprule
\textbf{Top Feat.} & \textbf{Adult} & \textbf{Child} & \textbf{Adolescent} &
\textbf{Clinical observation based on ~\cite{thabtah2017}} \\
\midrule
A9  & 0.719 & $0.240^\star$ & 0.058 & Social chit-chat  -- dominant child; drops in adolescents \\
A5  & $0.722^\star$ & 0.111 & $0.138^\star$ & Pattern recognition --- top adult \& adolescent feature \\
A4  & 0.708 & 0.180 & 0.083 & Routine disruption --- strong child; weaker in adolescents \\
A7  & 0.666 & 0.154 & 0.118 & Theory-of-mind (reading intentions) \\
A8  & 0.652 & 0.151 & 0.092 & Face-reading  \\
A3  & 0.618 & 0.037 & 0.026 & Lowest importance across all cohorts \\
\bottomrule
\end{tabular}
\end{table}

\paragraph{Adults.} Flat importance profile (A5: 0.722 $\approx$ A9: 0.719
$\approx$ A1: 0.716). No single feature dominates; 8 different features ranked
top across 17 models, reflecting multi-faceted adult ASD diagnostic patterns.

\paragraph{Children.} A9 (\emph{``I enjoy social chit-chat''}, reverse-scored)
strongly dominates (consensus importance 0.240), ranked top by 11 of 17 models.
A4 (0.180) and A7 (0.154) are secondary. Social motivation is the primary
early-childhood ASD marker.

\paragraph{Adolescents.} A complete feature hierarchy shift occurs: A5
(\emph{``I notice patterns in things all the time''}) leads (0.138), followed by
A7 (0.118) and A10 (0.098). A9 drops to 8th place (0.058), a stark contrast to
children. A5 wins the top-feature vote in 5 of 17 models; A6, A7, and A10 each
win 3, and the remaining 3 models select different top features, reflecting a
fragmented signal across all 17 models. This shift from social (A9) to
cognitive-perceptual (A5/A7) features is consistent with adolescent social
masking as a plausible hypothesis rather than a direct psychological measurement.

\subsection{Robustness}
\label{sec:robust}

\begin{figure}[H]
  \centering
  \includegraphics[width=\textwidth]{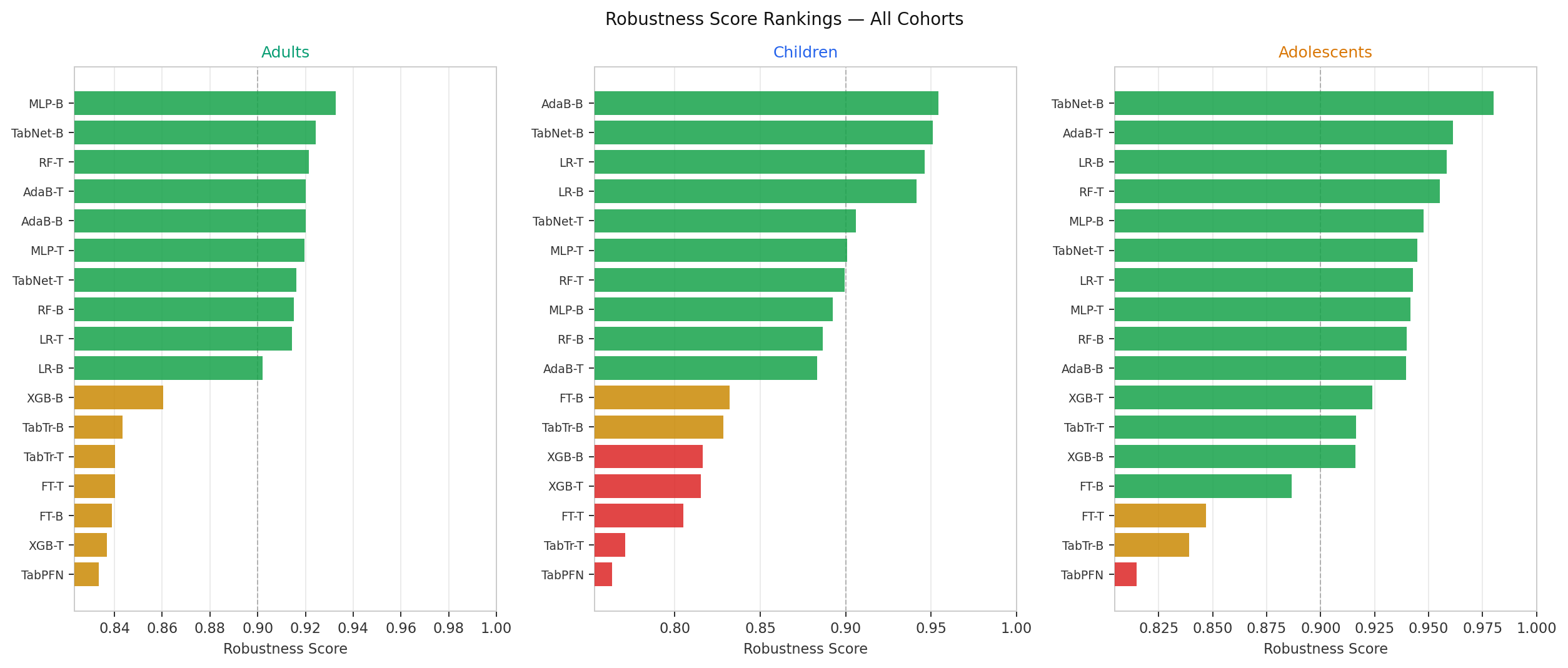}
  \caption{Robustness scores ranked by cohort.
           \textcolor{adultcol}{Green}: $\geq$0.88 (high);
           yellow: 0.82--0.88;
           \textcolor{warnred}{red}: $<$0.82.
           Dashed line: score\,=\,0.90.}
  \label{fig:rob}
\end{figure}

\begin{figure}[H]
  \centering
  \includegraphics[width=\textwidth]{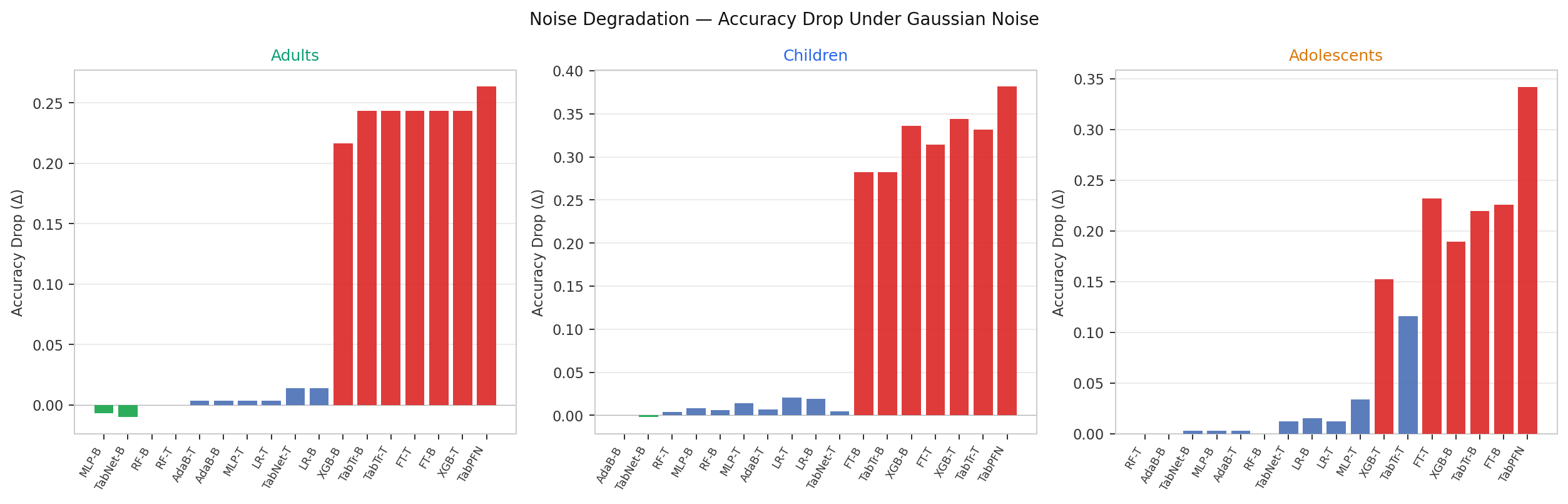}
  \caption{Average accuracy drop under Gaussian noise injection.
           Negative values indicate noise-immune models (slight regularisation
           benefit). Transformer models degrade by about 24\%; TabPFN shows
           a larger drop.}
  \label{fig:noise}
\end{figure}

\begin{figure}[H]
  \centering
  \includegraphics[width=\textwidth]{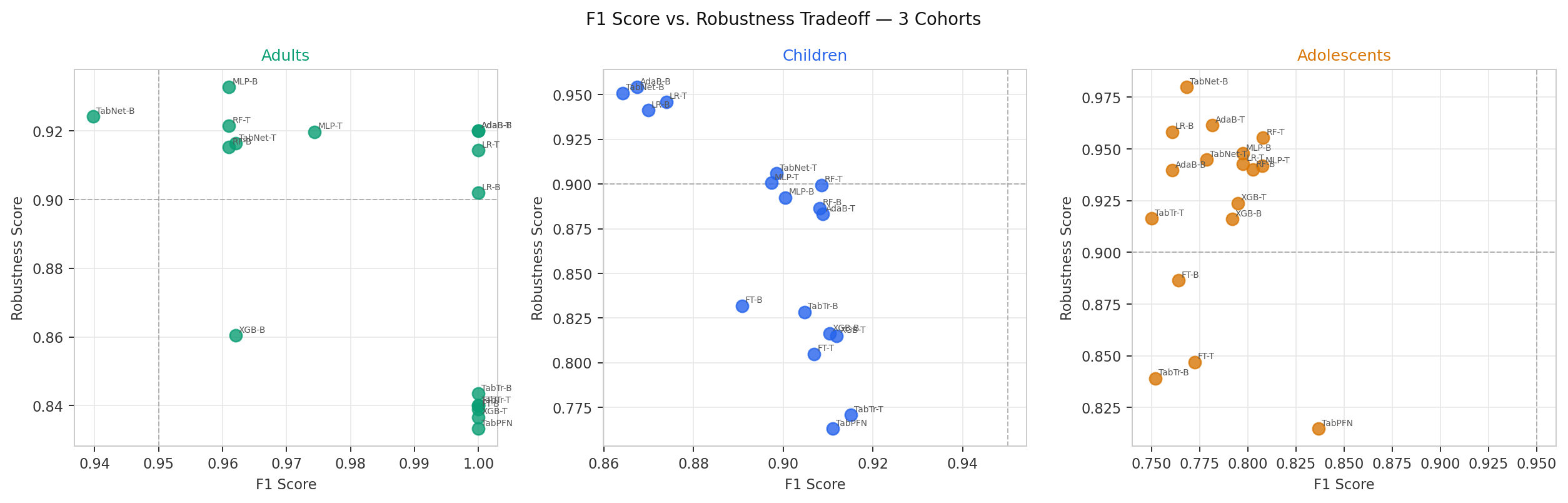}
  \caption{F1 Score vs.\ Robustness tradeoff per cohort.
           Dashed lines: F1\,=\,0.95 and Robustness\,=\,0.90.
           Ideal models appear top-right. The accuracy-robustness tradeoff is
           evident across all three cohorts.}
  \label{fig:tradeoff}
\end{figure}

\begin{table}[H]
\centering
\caption{Robustness scores and Gaussian noise degradation ($\Delta$acc)
         across all three cohorts.
         $\star$\,=\,cohort best robustness;
         $\dagger$\,=\,R\,$<$\,0.82 (critical risk).
         Negative noise $\Delta$ indicates noise-immune behaviour.
         Nine mid-tier models omitted; full results in supplementary material.}
\label{tab:rob}
\renewcommand{\arraystretch}{1.2}
\small
\begin{tabular}{L{3.0cm} C{1.5cm} C{1.5cm} C{1.5cm} C{1.5cm} C{1.5cm} C{1.5cm}}
\toprule
\textbf{Model}
  & \textbf{Rob.\ Ad.}
  & \textbf{Rob.\ Ch.}
  & \textbf{Rob.\ Ado.}
  & \textbf{Noise $\Delta$ Ad.}
  & \textbf{Noise $\Delta$ Ch.}
  & \textbf{Noise $\Delta$ Ado.} \\

\midrule
\multicolumn{7}{l}{%
  \small\textbf{Panel A\,---\,Robustness-Viable
  Models (strong robustness; noise sensitivity)}} \\
\midrule
MLP Baseline
  & $0.933^{\star}$ & 0.893 & 0.948
  & $-0.007$ & $+0.008$ & $+0.003$ \\
TabNet Baseline
  & 0.924 & 0.951 & $0.980^{\star}$
  & $-0.010$ & $-0.002$ & $+0.003$ \\
RF Baseline
  & 0.915 & 0.887 & 0.940
  & $+0.000$ & $+0.006$ & $+0.000$ \\
RF Tuned
  & 0.922 & 0.900 & 0.955
  & $+0.000$ & $+0.004$ & $+0.000$ \\
TabTransformer Tuned
  & $0.840$ & $0.771$ & 0.916
  & $+0.243$ & $+0.332$ & $+0.116$ \\
XGBoost Tuned
  & $0.837$ & $0.815$ & 0.924
  & $+0.243$ & $+0.344$ & $+0.152$ \\
TabPFN v2
  & $0.833$
  & $0.763^{\dagger}$
  & $0.815$
  & $+0.264$ & $+0.382$ & $+0.341$ \\

\midrule
\multicolumn{7}{l}{%
  \small\textbf{Panel B\,---\,Not
  Viable: robustness deficits compound weaknesses across other axes}} \\
\midrule
AdaBoost Baseline
  & 0.920 & $0.954$ & 0.940
  & $+0.003$ & $+0.000$ & $+0.000$ \\
FT-Transformer Tuned
  & $0.840$ & $0.805$ & 0.847
  & $+0.243$ & $+0.314$ & $+0.232$ \\

\bottomrule
\end{tabular}
\end{table}

The thresholds ($\geq$0.88 green, 0.82--0.88 yellow, $<$0.82 red) are 
empirically grounded in the bimodal distribution of noise degradation: 
models scoring $\geq$0.88 exhibit average noise degradation below 0.02, 
while those below 0.82 exceed 0.20---an order-of-magnitude separation 
reflecting the Lipschitz-continuity gap between piecewise-constant tree 
ensembles and deep transformers, whose compositional non-linearities 
amplify perturbations multiplicatively with depth~\citep{szegedy2014intriguing,
goodfellow2015explaining}, with the intermediate band capturing mixed 
degradation profiles warranting case-by-case review. The 0.90 dashed 
line marks the overfitting--robustness boundary made visible by the 
adult cohort ($n=148$), where models achieving perfect baseline accuracy 
(1.000) still fall into the yellow/red zone, indicating brittle decision 
boundaries rather than genuine generalisation~\citep{zhang2017understanding,
grinsztajn2022why}. Table~\ref{tab:rob} partitions models via cross-axis 
viability rather than robustness alone, since single-metric ranking 
obscures the compensatory structure governing clinical deployability: 
Panel~\ref{tab:rob}--A contains robustness-viable models whose noise 
sensitivity is externally manageable, evaluated \emph{jointly} with F1 
and ECE so that moderate robustness with strong calibration remains 
deployable (e.g., XGBoost Tuned: $R = 0.837$, ECE $= 0.022$, F1 $= 0.912$); 
Panel~\ref{tab:rob}--B excludes models on compound-axis grounds, with 
AdaBoost Baseline disqualified despite competitive robustness 
(0.920/0.954/0.940) by critical miscalibration in every cohort 
(ECE $= 0.302/0.190/0.157$, exceeding the 0.05 clinical 
threshold~\citep{vanCalster2019calibration}), and FT-Transformer Tuned 
excluded as Pareto-dominated by Panel~A alternatives with no compensating 
advantage on any remaining axis. Nine mid-tier models are omitted for 
brevity; full scores appear in the supplementary material.

Transformer-family noise brittleness (${\sim}24\%$ degradation under Gaussian perturbation) can be addressed without retraining via established clinical AI hardening techniques: input-level adversarial training and certified smoothing~\cite{cohen2019} improve effective robustness, while Monte Carlo Dropout and deep ensembles~\cite{lakshminarayanan2017} enable uncertainty-aware inference with automatic flagging of low-confidence cases for human-in-the-loop review, consistent with FDA SaMD guidance~\cite{fda2021}. Conformal risk control~\cite{angelopoulos2023} further provides distribution-free coverage guarantees valuable for the small-sample adolescent cohort. Population-level drift can be detected through AI observatory tools that trigger recalibration when incoming distributions diverge from training, a realistic concern in multi-site clinical rollout.

\subsection{HAP Metric Results}
\label{sec:hap_results}

\begin{figure}[H]
  \centering
  \includegraphics[width=\textwidth]{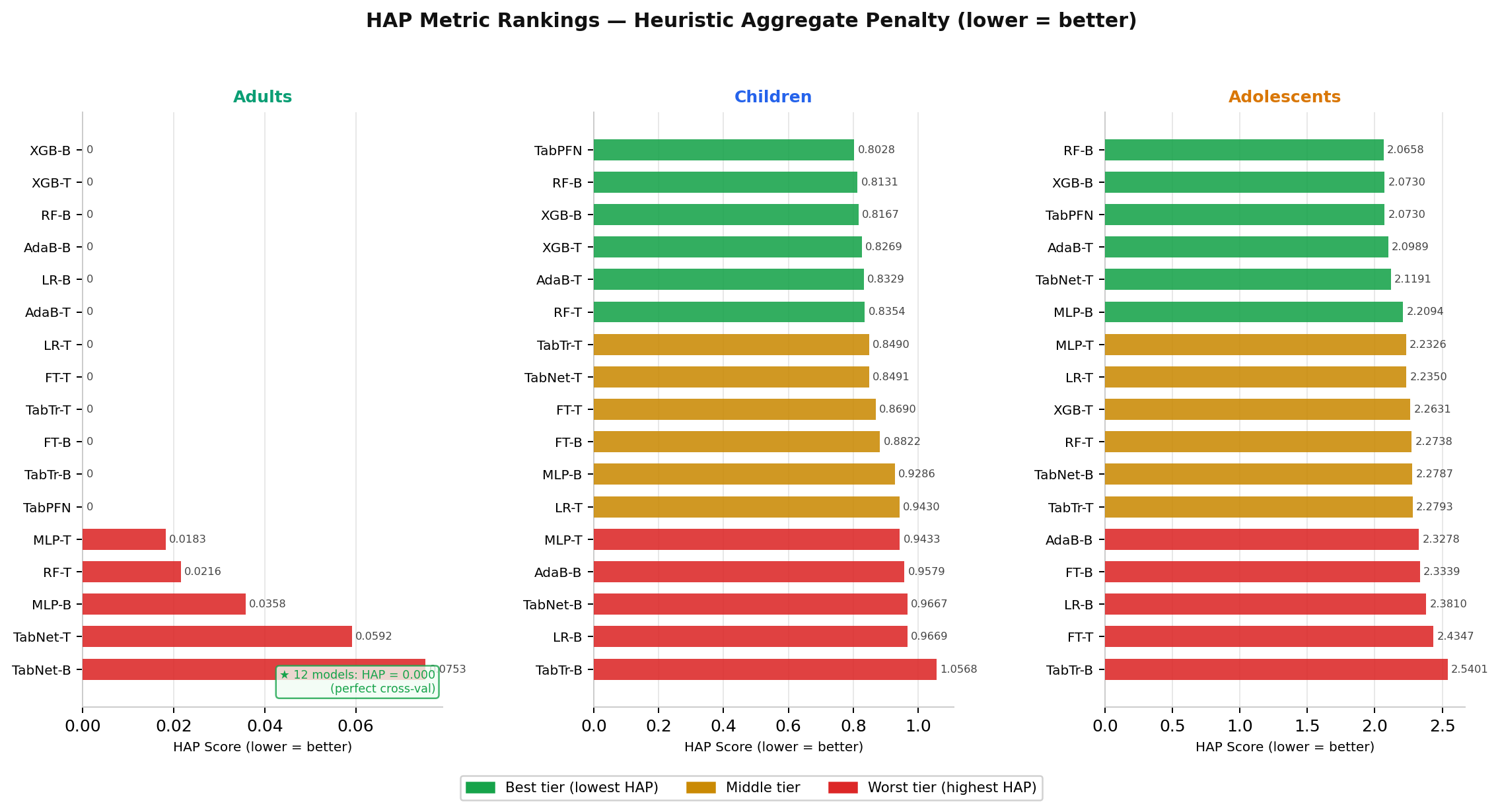}
  \caption{\hap{} metric rankings per cohort (lower\,=\,better), computed via
           5-fold stratified cross-validation with $\wfn=10$, $\wfp=2$,
           $\lambda=1.0$. Colours indicate performance tiers: green (lowest
           \hap{}), yellow (middle), red (highest \hap{}).}
  \label{fig:hap}
\end{figure}

\paragraph{Adults.}
Twelve of 17 models achieve \hap{}\,=\,0.000, meaning zero weighted
misclassifications across all five cross-validation folds, consistent with the
near-perfect F1 and AUC results on this cohort. The five non-zero models are
ordered: MLP Tuned (0.018), RF Tuned (0.022), MLP Baseline (0.036), TabNet
Tuned (0.059), and TabNet Baseline (0.075). These are driven primarily by
residual false negatives in harder folds rather than false positives, given the
$\wfn:\wfp = 5:1$ penalty structure.

\paragraph{Children.}
\hap{} rankings diverge meaningfully from F1 rankings. TabPFN leads
(0.803) despite ranking only 7th on F1 (0.911), reflecting its superior
probability calibration minimising costly false negatives. TabTransformer Baseline
scores worst (1.057); its moderate ECE (0.044) compounds FN errors into elevated
cross-fold variance. AdaBoost Baseline ranks 14th on \hap{} (0.958) despite
competitive F1, penalised for its severe miscalibration (ECE\,=\,0.190) inflating
the variance term $\lambda \cdot \mathrm{Var}$.

\paragraph{Adolescents.}
\hap{} scores are scaled roughly $2.5\times$ higher than children, directly
reflecting the lower absolute F1 (more FNs per fold). RF Baseline, XGBoost
Baseline, and TabPFN tie for best (2.073), the same three models
that lead on raw AUC. TabTransformer Baseline scores worst (2.540), consistent
with its poor adolescent F1 (0.752) and moderate ECE (0.082). The \hap{} ranking
here aligns closely with F1 rankings, suggesting that at lower performance levels,
false-negative volume dominates over variance as the primary cost driver.

\subsection{Four-Axis Summary Scorecard}
\label{sec:scorecard}

\begin{figure}[H]
  \centering
  \includegraphics[width=0.88\textwidth]{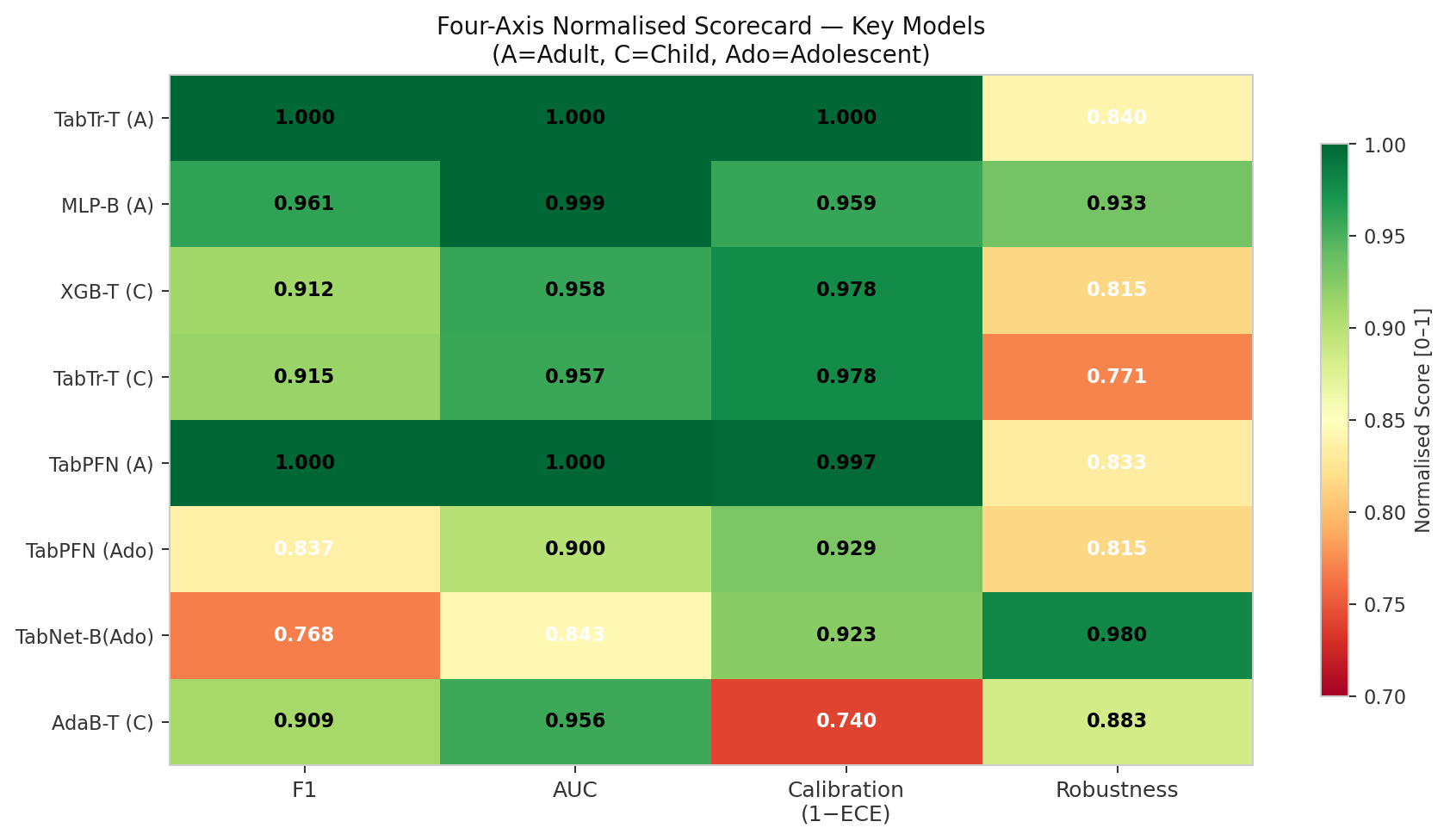}
  \caption{Four-axis normalised scorecard for eight key models
           (A\,=\,Adult, C\,=\,Child, Ado\,=\,Adolescent).
           All axes normalised to $[0,1]$; higher\,=\,better.
           Calibration plotted as $1 - \mathrm{ECE}$.}
  \label{fig:scorecard}
\end{figure}

\noindent
The radar profiles below highlight the recommended models per cohort.

\begin{figure}[H]
  \centering
  \includegraphics[width=0.52\textwidth]{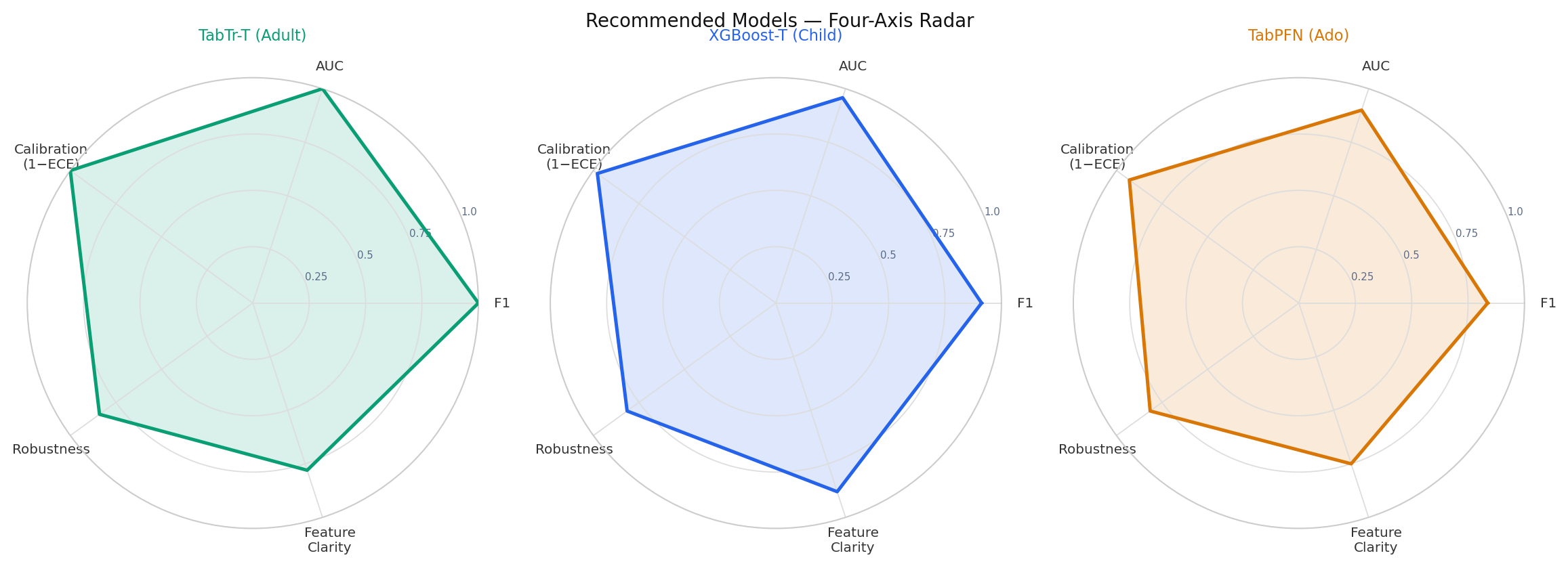}
  \caption{Four-axis radar profiles for the three recommended models:
           \textcolor{adultcol}{\textbf{TabTransformer Tuned}} (adult),
           \textcolor{childcol}{\textbf{XGBoost Tuned}} (child),
           \textcolor{adocol}{\textbf{TabPFN~v2}} (adolescent).
           Axes: F1, AUC, Calibration ($1-\mathrm{ECE}$), Robustness, feature clarity}
  \label{fig:radar}
\end{figure}

\section{Discussion}
\label{sec:discussion}

Building on the quantitative results presented above, we now analyse the qualitative factors driving model performance and their clinical implications.

\subsection{Child--Adolescent Performance Gap}

Children and adolescents are genuinely distinct cohorts with different
classification difficulty and feature hierarchies. The adolescent F1 ceiling is
0.837 (TabPFN) vs 0.915 (TabTransformer Tuned) for children, a 7.8
percentage-point gap that
persists across all 17 models. Adolescent AUC similarly peaks at 0.900 vs 0.963
for children. Beyond performance, the feature hierarchy shifts completely: A9 (social
motivation) dominates children while A5 (pattern recognition) leads adolescents,
with A9 dropping to 8th place. This is clinically interpretable as social masking:
adolescents learn to compensate surface social behaviours while perceptual
rigidities persist. These findings argue strongly for cohort-specific models
in clinical ASD screening rather than age-agnostic deployment.
A larger, clinically verified adolescent-specific dataset is needed to
maximise predictive performance.

\subsection{Accuracy--Robustness--Calibration Tradeoffs}

A recurring pattern is strong dissociation across evaluation axes. Transformer
architectures achieve perfect F1 on adults but rank among the least robust
models ($R$\,=\,0.833--0.840) with around 24\% noise-induced accuracy degradation.
AdaBoost achieves F1\,=\,1.000 with ECE\,=\,0.302. TabPFN achieves best AUC
with worst robustness, which can be countered using external hardening methods. Adolescents exhibit an unexpected pattern: models are
\emph{uniformly more robust} than on children (TabNet Baseline: 0.980 ado
vs 0.951 child; even TabPFN last at 0.815 outperforms most child-cohort scores)
despite lower absolute F1, suggesting a smoother decision boundary under
perturbation. For less robust models, established external hardening techniques like conformal prediction can be applied to mitigate robustness deficits.

\begin{table}[H]
\centering
\caption{Deployment recommendations by cohort and setting, with supporting metrics
         drawn from Tables~\ref{tab:f1_auc}, \ref{tab:ece_a}, \ref{tab:brier_b},
         and \ref{tab:rob}.
         $^\dagger$Pair with uncertainty flagging for $p \in [0.40,\,0.60]$.}
\label{tab:deployment}
\renewcommand{\arraystretch}{1.4}
\setlength{\tabcolsep}{5pt}
\small
\begin{tabular}{L{2.0cm} L{2.2cm} L{3.0cm} C{1.2cm} C{1.2cm} C{1.2cm} C{1.4cm}}
\toprule
\textbf{Cohort} & \textbf{Setting} & \textbf{Recommended Model} &
\textbf{F1} & \textbf{AUC} & \textbf{ECE} & \textbf{Robust.} \\
\midrule
\multirow{2}{*}{\textbf{Adult}}
  & Controlled
  & \textbf{TabTransformer Tuned}
  & 1.000 & 1.000 & $2.1{\times}10^{-7}$ & 0.840 \\
  & Noisy
  & \textbf{MLP Baseline}
  & 0.961 & 0.999 & 0.041 & \textbf{0.933} \\
\midrule
\multirow{2}{*}{\textbf{Child}}
  & General
  & \textbf{XGBoost Tuned}
  & 0.912 & 0.958 & 0.022 & 0.815 \\
  & Noisy
  & \textbf{RF Tuned}
  & 0.909 & 0.954 & 0.037 & \textbf{0.900} \\
\midrule
\multirow{2}{*}{\textbf{Adolescent}}
  & General
  & \textbf{TabPFN v2}$^{\dagger}$
  & \textbf{0.837} & \textbf{0.900} & 0.071 & 0.815 \\
  & Noisy
  & \textbf{TabNet Baseline}
  & 0.768 & 0.843 & 0.077 & \textbf{0.980} \\
\bottomrule
\end{tabular}
\end{table}

\subsection{AdaBoost Miscalibration}

AdaBoost's critical miscalibration (ECE\,$\approx$\,0.190--0.303 across cohorts)
despite competitive F1 is well-explained by boosting's confidence amplification.
Any clinical deployment must apply mandatory post-hoc calibration. We recommend
treating AdaBoost F1 results as upper bounds on operational performance.

\subsection{TabPFN as a Clinical Foundation Model}

The comparison between TabPFN~v2 and fully tuned models is deliberately
asymmetric: TabPFN uses a fixed pre-trained network with in-context
learning (no gradient updates or hyperparameter search on the target
data) while all other models receive extensive optimisation. Despite
this disadvantage, TabPFN achieves the highest child AUC (0.963) and
competitive child F1 (0.911). Its calibration is
consistently strong (ECE\,$\approx$\,0.003--0.071 across cohorts). A
fine-tuned TabPFN would likely close or eliminate the remaining F1 gap;
this comparison is planned as future work. The critical limitation is
robustness (ranked last in all cohorts). We recommend pairing TabPFN
with uncertainty monitoring and external robustness-enhancement
techniques: predictions with $p \in [0.40, 0.60]$ should be
automatically routed to human clinical review.

\subsection{HAP in Clinical Context}

The penalty ratio $\wfn:\wfp = 5:1$ reflects a straightforward clinical
reality: missing a genuine ASD case costs far more than triggering an
unnecessary follow-up referral. Deployment teams can adjust this ratio to
fit their setting---higher $\wfp$ in resource-constrained environments,
higher $\wfn$ in universal screening programmes---without changing which
models come out ahead.

The variance penalty $\lambda = 1.0$ serves an equally practical purpose:
it down-ranks models that are accurate on average but inconsistent across
sites or patient subgroups, since a model that behaves erratically across
data partitions is a deployment risk regardless of its headline score.

Beyond evaluation, \hap{} admits a useful geometric interpretation: it maps
the misclassification space $(\mathrm{FN}, \mathrm{FP})$ which are independent of the model weights and hyperparameters into a single
discriminative score, tracing a trajectory through model space as $\lambda$
varies. This structure makes \hap{} a natural candidate for an additional inverse reward
signal in reinforcement learning settings, where the negative \hap{} score
can guide a policy toward the optimal operating point in the
$(\mathrm{FN}, \mathrm{FP})$ plane rather than optimising a flat accuracy
surface. The $\lambda$-parameterised trajectory then acts as a curriculum,
progressively penalising instability as training matures.

In short, \hap{} captures two properties that standard metrics miss: what
\emph{kind} of error a model makes, and how \emph{consistently} it behaves.
Its strong agreement with the 4-axis analysis recommendations suggests it
reflects genuine clinical utility rather than a statistical artefact. We
recommend reporting \hap{} alongside F1 and AUC in future clinical AI
benchmarks.

\section{Limitations}
\label{sec:limitations}

\subsection{Dataset and Cohort Constraints}

The v3 dataset relies entirely on self-reported AQ-10 responses collected
through a single application (ASDTest), raising concerns about response bias
and limited generalisability to broader clinical populations. The adult
cross-dataset duplication rate (54.0\%) undermines the independence
of the two combined sources: after deduplication only 148 unique adult
records remain, yielding a test set too small to reliably distinguish
genuine model capability from dataset simplicity or residual distributional
overlap. The near-perfect adult scores (10/17 models at F1\,=\,1.000)
should therefore be interpreted with caution---they likely reflect the
constrained sample size and limited feature diversity rather than true
diagnostic task saturation. The adolescent cohort (818 records)
remains small; the authors
acknowledge that a larger, clinically verified adolescent-specific dataset
is needed to meaningfully raise the F1 ceiling beyond 0.837. Furthermore,
the exclusion of demographic variables (gender, ethnicity, country of
residence) and the absence of a stratified fairness analysis limit the
clinical completeness of the evaluation. Although the dataset contains
meaningful demographic diversity (67.6\% male; five ethnic groups
represented), the per-subgroup sample sizes are too small to support
statistically meaningful stratified performance comparisons (e.g., the
Black subgroup comprises only 4.1\% of records). A properly powered
fairness evaluation would require targeted oversampling or a larger,
demographically balanced collection effort.

\subsection{Evaluation and Metric Limitations}

A fundamental limitation is the absence of multi-site or external
validation: all models are trained and evaluated on data drawn from a
single-source distribution (the ASDTest application). No independent
clinical site, alternative screening instrument, or geographically
distinct population was used to verify that the reported performance
transfers beyond the originating data collection context. Consequently,
all results should be treated as internal benchmark scores rather than
estimates of real-world diagnostic performance.

Additionally, all predictive performance metrics (F1, AUC) are reported on a single
stratified train--test split without repeated random seeds or statistical
significance testing; observed differences between models may therefore
reflect split-specific variation rather than true performance gaps. The
\hap{} metric, while clinically motivated, employs penalty weights that are acknowledged to be
statistically approximated rather than empirically elicited from domain experts, limiting
its authority as a standardised clinical measure. Additionally, the comparison between fully hyperparameter-tuned
models and TabPFN~v2 (which performs in-context learning on a fixed
pre-trained network without any gradient updates or hyperparameter
search on the target data) introduces an inherent asymmetry in the
evaluation design that should be interpreted with caution.

\subsection{Clinical Validity}

The AQ-10 is a screening instrument rather
than a diagnostic tool, and all benchmark evaluations validate model
predictions against questionnaire scores rather than formal clinician-confirmed
ASD diagnoses. Consequently, even a model achieving perfect F1 on the
held-out test set provides no guarantee of real diagnostic utility in a
clinical pathway. Until prospective, multi-site validation against
clinician-confirmed diagnoses is conducted across diverse healthcare
settings, the findings presented in this work should be interpreted as
proof-of-concept evidence rather than a demonstration of clinical
deployability.

\section{Conclusion}
\label{sec:conclusion}

We presented a four-axis benchmark of 17 ML/DL/foundation models for
ASD questionnaire screening across child, adolescent, and adult
cohorts. All results are derived from a single-source distribution with
no multi-site or external validation; they represent internal benchmark
performance and should not be interpreted as clinical generalisability
estimates.

\paragraph{Core findings.}
The adult cohort achieves near-perfect held-out scores (10 of 17 models
at F1\,=\,1.000), though this likely reflects the small sample size
($n$\,=\,148 after deduplication) and limited data diversity rather than
genuine diagnostic task saturation. Cohort-specific
feature hierarchies revealed developmental shifts in ASD phenotypic
expression. A9 (social motivation) emerged as the dominant feature for
children (importance 0.240, ranking top in 11 of 17 models), while A5
(pattern recognition) led for adolescents (0.138, with a fragmented signal
spread across A5, A6, A7, and A10). Adults displayed a flat multi-feature
profile with no single dominant predictor.

\paragraph{Adolescent cohort difficulty.}
The adolescent cohort is a harder classification task, with an F1 ceiling of 0.837 compared to 0.915 for children. However, adolescents show higher robustness (TabNet Baseline: 0.980), suggesting a smoother decision boundary under perturbation. These results indicate that age-agnostic models are inadvisable, but they remain benchmark-level evidence rather than proof of clinical diagnostic readiness.

\paragraph{HAP contribution.}
The proposed HAP framework (with tunable parameters) offers a principled clinical penalty structure that complements standard metrics, with potential application for model optimisation through reinforcement learning.

\paragraph{Deployment recommendations.}
Model recommendations vary by cohort and deployment context: TabTransformer Tuned for adults in controlled settings, MLP Baseline for adults in noisy environments, XGBoost Tuned for children, TabPFN v2 for adolescents where accuracy is the priority, and TabNet Baseline for adolescents where robustness is prioritised. These deployment recommendations should still be interpreted cautiously because no stratified fairness analysis was run during model selection.

\paragraph{Future work.}
Future directions include prospective multi-site validation on
independent clinical populations, a stratified fairness analysis across
gender and ethnicity subgroups (requiring a larger, demographically
balanced dataset), age-adaptive ensemble systems, and formal HAP weight
elicitation from domain experts. On the modelling side, we plan a
comparative study of tabular foundation models including a fine-tuned
TabPFN across three cohorts, working toward a medical foundation model
for ASD and related neurodevelopmental conditions. We further aim to
apply symbolic regression and mechanistic interpretability to extract
clinically transparent ASD concepts from learned representations.
Finally, the hard asymmetric HAP penalty will be replaced by a
confidence-weighted variant that focuses learning on high-confidence
false negatives rather than penalising all missed positives uniformly.


\section*{Data and Code Availability}
\label{sec:availability}
The ASD-Bench v3 dataset (4{,}068 records across three age cohorts) is publicly available at \url{https://huggingface.co/datasets/rescommons/ASD_V3_all_ages}. The complete evaluation pipeline will be released upon acceptance.

\section*{Ethics Statement}

This study uses publicly available, de-identified questionnaire data from the
UCI Machine Learning Repository and a supplementary source. No new human
subjects data were collected. All analyses were conducted on pre-existing,
anonymised records. The AQ-10 is a screening instrument, not a diagnostic
tool; model outputs should not be used as a substitute for clinical assessment.

\bibliographystyle{plainnat}
\bibliography{references}

\end{document}